\documentclass{article}

\usepackage[english]{babel}
\usepackage[square,numbers]{natbib}
\usepackage[letterpaper,top=2cm,bottom=2cm,left=3cm,right=3cm,marginparwidth=1.75cm]{geometry}

\usepackage[export]{adjustbox}
\usepackage{amsmath}
\usepackage{graphicx}
\usepackage{caption}
\usepackage[colorlinks=true, allcolors=blue]{hyperref}
\usepackage{authblk}
\usepackage{float}
\usepackage{subcaption}

\title{\textbf{Universal Lesion Segmentation Challenge 2023: \\ A Comparative Research of Different Algorithms}}
\author[1]{Kaiwen Shi}
\author[2]{Yifei Li}
\author[1]{Binh Ho}
\author[2]{Jovian Wang}
\author[2]{Kobe Guo}
\affil[1]{Mathematics Department, Vanderbilt University, Nashville, TN, USA}
\affil[2]{Department of Computer Science, Vanderbilt University, Nashville, TN, USA}
\date{March 30, 2024}

\begin{document}
\maketitle

\begin{abstract}
    In recent years, machine learning algorithms have achieved much success in segmenting lesions across various tissues. There is, however, not one satisfying model that works well on all tissue types universally. In response to this need, we attempt to train a model that 1) works well on all tissue types, and 2) is capable of still performing fast inferences. To this end, we design our architectures, test multiple existing architectures, compare their results, and settle upon SwinUnet. We document our rationales, successes, and failures. Finally, we propose some further directions that we think are worth exploring. codes: https://github.com/KWFredShi/ULS2023NGKD.git
    \\
\end{abstract}

\section{Introduction}
\hspace{0.5cm} Medical image segmentation is a crucial task in medical image processing. Thanks to the advent of CNN\cite{NIPS1989_53c3bce6}, U-Net \cite{RonnebergerFB15}, and their variants such as V-Net\cite{7785132}, 3D U-Net\cite{DBLP:journals/corr/CicekALBR16}, Res-UNet\cite{Xiao2018WeightedRF}, Dense-UNet\cite{8379359}, we are able to perform segmentation task with precision.  More recently, with implementations of transformer-based models, the medical imaging community enjoyed satisfying success in segmentation tasks. Networks like Medical Transformers\cite{DBLP:journals/corr/abs-2102-10662medtrans} and SwinUnet\cite{cao2021swinunet} push the front-line boundary to another degree. Others have implemented learning methodologies from other fields, such as dictionary learning, to work on medical images. KEN\cite{10.1007/978-3-031-20074-8_28KEN} - knowledge embedding network - for example, takes advantage of the fruitfulness of information embedding in each layer via dictionary learning to provide a more semantically meaningful network.

While most of the networks have achieved very promising results on datasets composed of one tissue type, none of them works well universally across various tissue types. Meanwhile, facing a growing need for CT exams and their lesion segmentation, radiologists have suffered much from intensive human labor. In response to these two situations, Max de Grauw, Bram van Ginneken, and Alessa Hering of the Diagnostic Image Analysis Group, partnering with Mathai Tejas, Pritam Mukherjee, and Ronald Summers of the National Institutes of Health, launched the Universal Lesion Challenge 23 (https://uls23.grand-challenge.org/). 

To address the dire situation, we attempt to adopt/devise an algorithm that is 1) precise enough based on the Dice score, 2) adequately robust to respond to variation in tissue types, and 3) light-weight in model complexity so that the inference runs around 5 seconds per input. To this end, we test the effectiveness of various algorithms, including nnUNetv2\cite{Isensee2020nnUNetAS}, DeepLabV3+\cite{DBLP:journals/corr/ChenPSA17DLV3}, Medical Transformer\cite{DBLP:journals/corr/abs-2102-10662medtrans}, SwinUnet\cite{cao2021swinunet}, and TransUNet\cite{DBLP:journals/corr/abs-2102-04306TransUNet}. We compare their results and fine-tune our final model on TransUNet, the one that has worked best in our testing.

\textbf{Contributions.} 1. We tested the effectiveness of nnUNetv2, DeepLabV3+, Medical Transformer, SwinUnet, and TransUNet on the Bone Lesion dataset. 2. We fine-tuned the TransUNet model, with well-written data augmentations and transformations. 3. We included a discussion on potential improvements that can be made to our project.

\section{Task Description}

\hspace{0.5cm} The task at hand is to design, implement, and train a model that can segment lesions universally. To be more specific, the input of the model will be a 256x256x128 data point, with only one channel. The input is named VOI, or volume of interest. The output of the model should be a segmentation of the lesion, and the performance will be measured with inference speed, segmentation accuracy, and robustness. 
\begin{figure}[H]
    \centering
    \includegraphics[width = 0.7\textwidth]{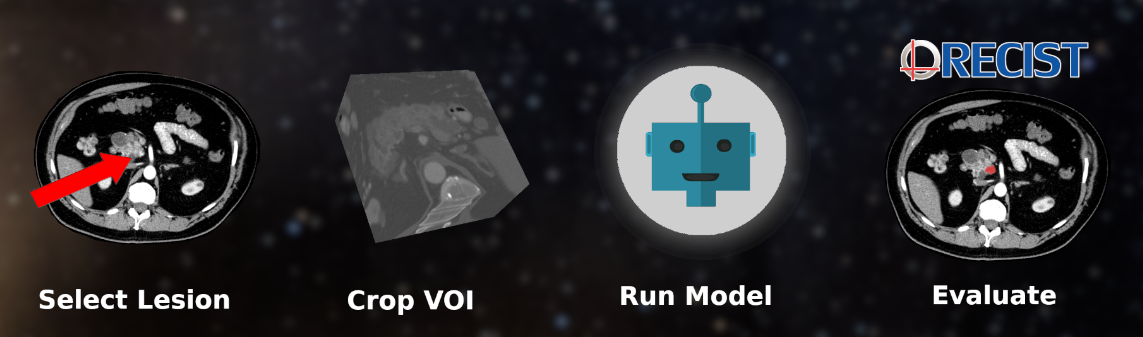}
    \caption{The Proposed Pipeline from ULS Challenge Host}
\end{figure}

\hspace{0.5cm} Once the model is trained and deployed, radiologists can locate the VOI manually, pass the VOI into the model pipeline, and get the returned lesion segmentation almost immediately.

\section{Experiments}
\hspace{0.5cm} In this section, we describe the rationale behind the use and test of these models, the methodologies and highlights of each model, and their corresponding results.
\subsection{Rationale}
\hspace{0.5cm} We started the project by testing the effectiveness of each model on the Bone dataset, Part 1. We split the data into a 0.7-0.2-0.1 train-test-validation compartmentalization. We trained the models on the train data and compared the validation scores. We understood that the model that worked well on the Bone dataset did not necessarily tell its effectiveness on the entire data population, so after having initial results that did not deviate too much, we further trained the working models on the whole Part 1 data. We again compared the results and went on with the most promising model to complete the project.

\subsection{Models and Results}
\subsubsection{nnUNetv2 - Baseline}

\hspace{0.5cm} nnUNetv2\cite{Isensee2020nnUNetAS} is an advanced neural network architecture designed for medical image segmentation tasks. Built upon the success of its predecessor, nnUNet, nnUNetv2 incorporates several enhancements and optimizations, making it a state-of-the-art solution in the field of medical image analysis.

\hspace{0.5cm} nnUNetv2 adopts a cascaded architecture, consisting of multiple processing stages, each refining the segmentation output progressively. The network architecture comprises deep convolutional neural networks (CNNs), augmented with advanced modules such as residual connections, dilated convolutions, and attention mechanisms. These components enable nnUNetv2 to capture complex spatial dependencies and contextual information crucial for accurate segmentation. Some of its highlights include performance, versatility, and robustness. 

\hspace{0.5cm} nnUNetv2 served as the baseline model in the ULS 23 challenge. It has produced very promising results on data points from one tissue type. However, when training on multi-typed data, the model exhibited a large variation across tissue types and even within each tissue type.
\begin{figure}[H]
    \centering
    \includegraphics[width=0.7\textwidth, center]{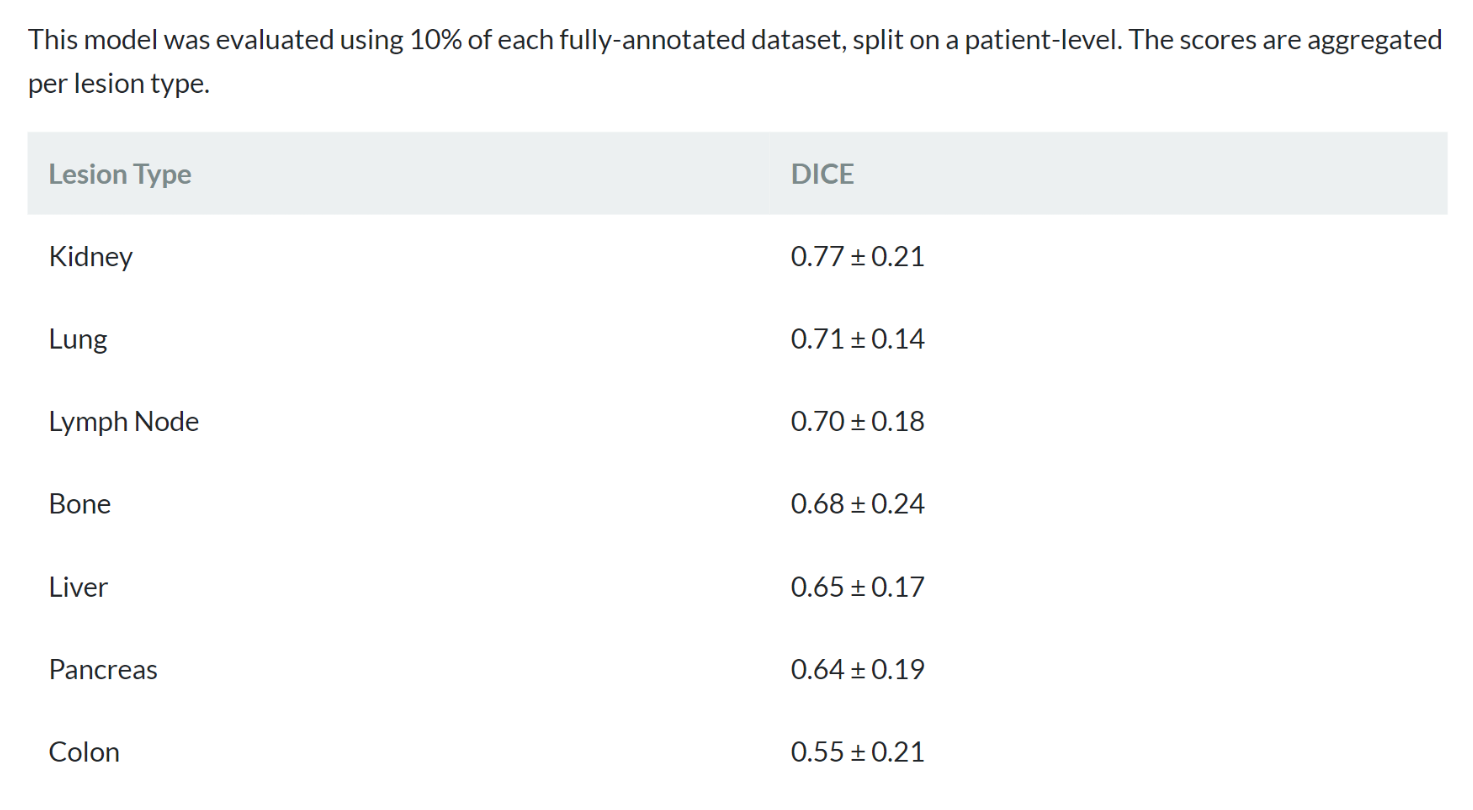}
    \caption{ Baseline Result}
\end{figure}

\hspace{0.5cm} Since the nnUNetv2 pipeline had written optimizers to tune its hyperparameters, we did not continue fine-tuning it. We instead use it as a benchmark against which we test our experiments.

\subsubsection{Medical Transformer}
\hspace{0.5cm} Medical Transformer\cite{DBLP:journals/corr/abs-2102-10662medtrans} utilized the LOGO learning strategy and Gated Axial Attention. The former strategy helps to discern the local features while also caring for the global context\cite{DBLP:journals/corr/abs-2010-11929Vit}, and the latter Attention mechanism saves time complexity while giving more attention to the axial components.\cite{DBLP:journals/corr/abs-1912-12180AA}
\begin{figure}[H]
    \centering
    \includegraphics[width=0.7\textwidth, center]{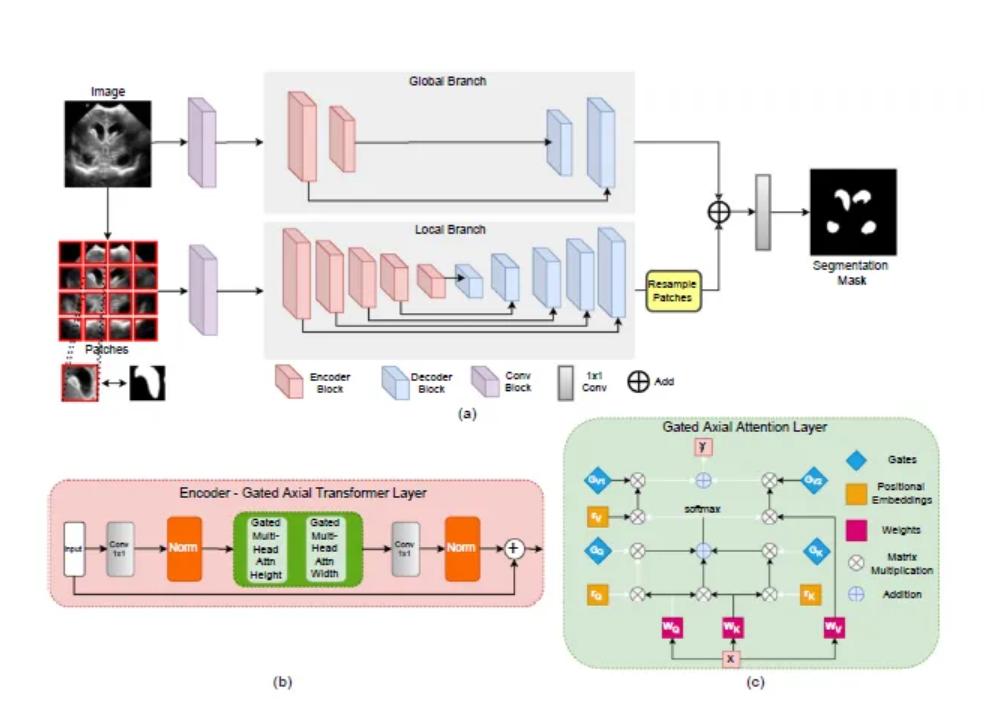}
    \caption{ LOGO Learning and Gated Axial Attention}
\end{figure}
The structure of Medical Transformer gave it prime effectiveness in segmentation tasks on MoNuSeG, GLAS, and Brain Anatomy US datasets. The problem of adopting this model, however, lay with its fixed input dimension and type. The input of the original Medical Transformer was a 128x128 image, while the VOI that was put into the model pipeline was a 256x256x128 volume. The disagreement with regard to the inputs called for modification of the existing model.

The team primarily experimented with three different potential fixes. The first one involved simple flattening of the VOI into 2D images and adjusting the model parameters according to the size of the image; the second one utilized an FCNN to extract the features from the input and fit the extracted feature into the model; the last one, being the most straightforward solution, was to implement a 3D version of Medical Transformer.

None of the fixes, unfortunately, worked well to address the current challenge at hand. The first two fixes, no matter where the feature extraction filters were put or simple flattening, took more than 100 GB of RAM to initialize the model. The last solution, even though it seemed more promising, suffered from the same problem of exploding model size. Therefore, the Medical Transformer failed to solve our problem.

\subsubsection{SwinUnet}

\hspace{0.5cm} The SwinUNet model\cite{cao2021swinunet} differs from UNet due to its use of the Swin transformer blocks. It uses a hierarchical Swin transformer with shifted windows as the encoder and a symmetric Swin transformer as the decoder. This allows the model to adeptly capture complex patterns in medical images. The hierarchical design lets the model efficiently process different scales of spatial context.
\begin{figure}[H]
    \centering
    \includegraphics[width=0.7\textwidth]{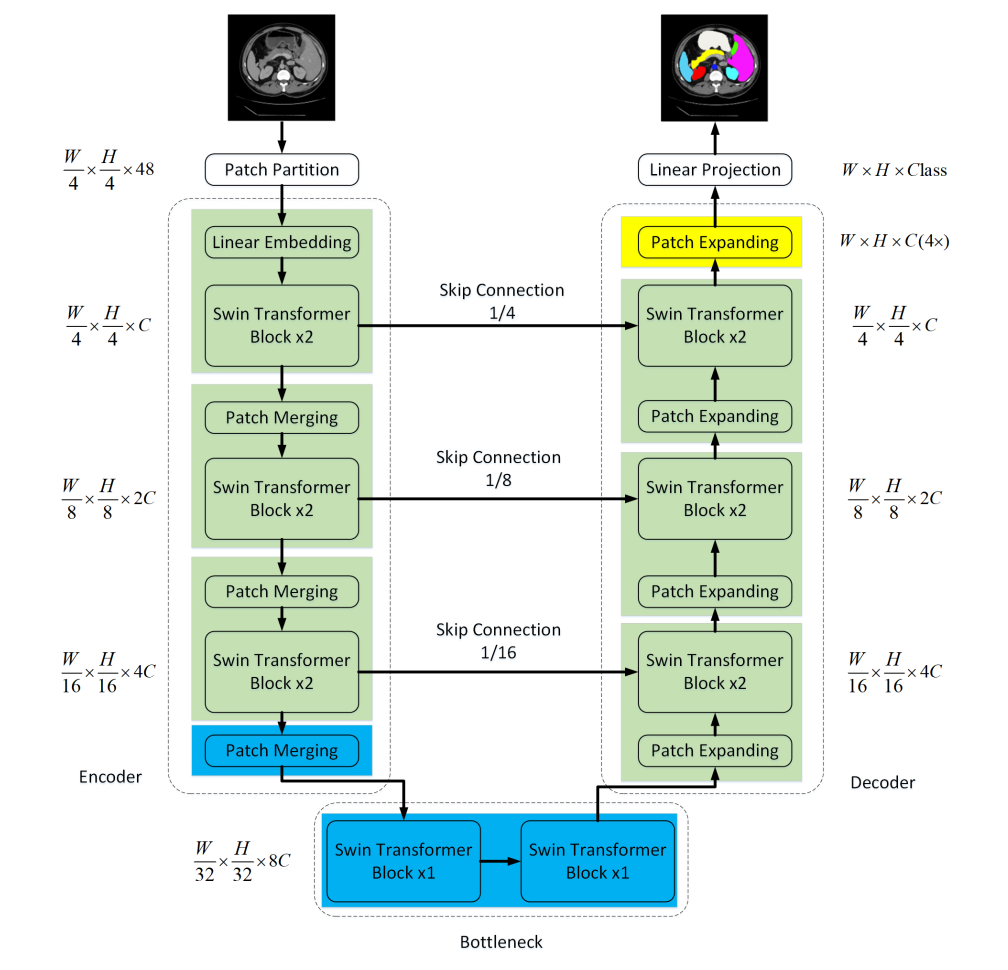}
    \caption{SwinUNet Architecture}
\end{figure}

\hspace{0.5cm}Our experimental process began with training the model on the ULS23 bone dataset. Through the training, we were able to track the model's training dice loss and validation dice loss metrics. The training dice loss had a consistent downward trend, which indicated an increase in dice score, but the validation dice loss had upward spikes throughout the iterations. This showed us that the model could have been overfitting on the dataset.
\begin{figure}[H]
    \centering
    \includegraphics[width=0.7\textwidth]{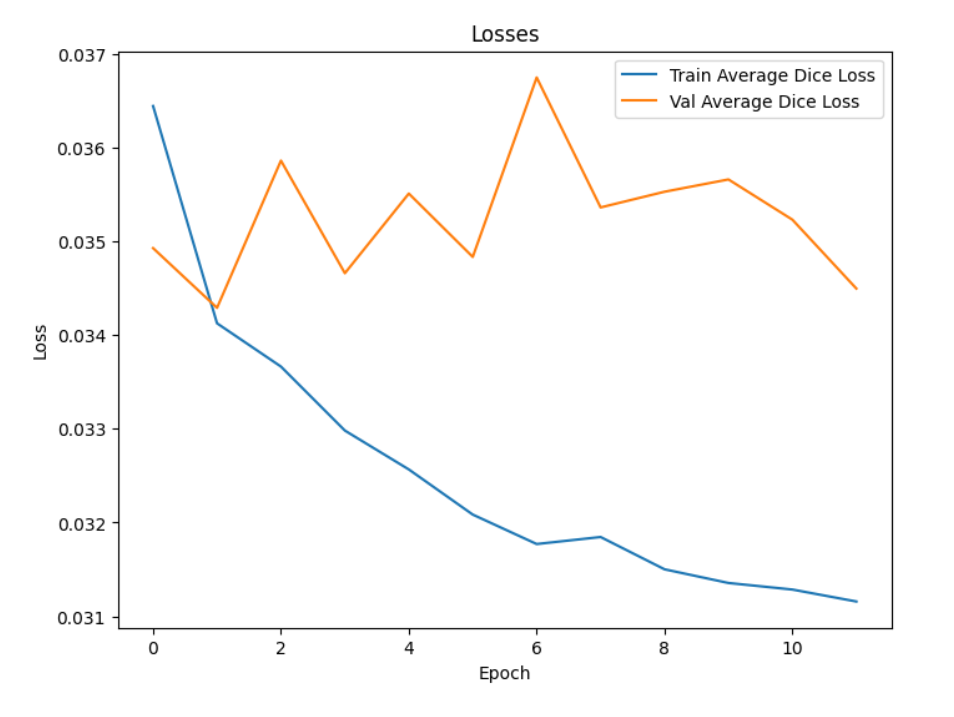}
    \caption{SwinUNet Result}
\end{figure}

\hspace{0.5cm}To mitigate overfitting, we employed data augmentation techniques such as random rotations, horizontal and vertical flips, and Gaussian blurs onto the medical images in the dataset. Even through this, however, the validation training loss did not improve much, which led to this model not being our best one.

\subsubsection{DeepLabV3+}

\hspace{0.5cm} The DeepLabV3+\cite{DBLP:journals/corr/abs-1802-02611DLV3+} is designed with a four-layer architecture featuring downsampling and upsampling blocks, integrated with skip connections to facilitate information flow. In the downsampling path, convolutional layers extract and learn features using convolutional filters followed by max pooling. The upsampling path then reconstructs the spatial dimensions through transposed convolution. Skip connections bridge these paths, allowing information to be directly passed from downsampling blocks to the corresponding upsampling blocks to mitigate the loss of finer details during the simplification of the image.
\begin{figure}[H]
    \centering
    \includegraphics[width=0.5\textwidth, height=0.3\textheight]{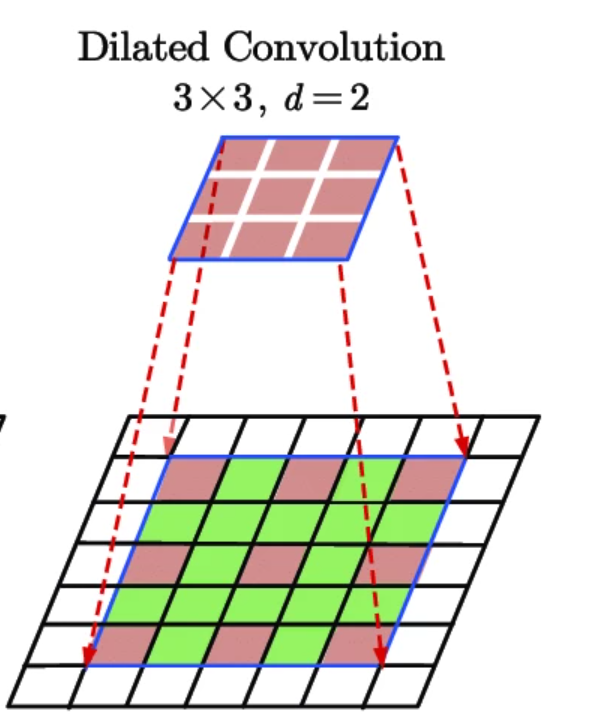}
    \caption{Atrous/Dilated Convolution}
\end{figure}

\hspace{0.5cm} We began the experiment with the 2D DeepLabV3\cite{DBLP:journals/corr/ChenPSA17DLV3} model. We flattened the input directly and sent it into the inference pipeline. Initially promising, this model displayed tendencies of overfitting, as evidenced by its attainment of a 0.82 Dice coefficient following 2500 epochs of training—surpassing the baseline model's 0.72 Dice score significantly on the bone dataset.

\hspace{0.5cm}In response, we ventured into a 3D U-Net model, referred to as DeepLabV3+, owing to its atrous convolution (or dilated convolution) features similar to the DeepLabV3 architecture. However, the complex structure of this 3D model introduced significant challenges, notably extending the training duration. To overcome the challenges, we tried various local pooling techniques before inferencing, but the result turned out bleak: After 25 epochs of training, the dice score was only around 0.1, which made this model nearly useless for our purpose.

\subsubsection{TransUNet}

\hspace{0.5cm} TransUNet\cite{DBLP:journals/corr/abs-2102-04306TransUNet} is a model that aims to improve the performance of the UNet architecture by leveraging Transformer blocks as the encoders for medical image segmentation tasks. In particular, while the UNet architecture is better than pure Transformers-based models in extracting low-level details, the UNet architecture alone is not sufficient in modeling long-range dependencies, a drawback that is directly addressed by the introduction of Transformer blocks. Therefore, by using Transformer blocks as the encoders in a UNet-like architecture, TransUNet is able to efficiently tackle both long-range dependencies as well as low-level details. In their original paper, the authors of TransUNet show that TransUNet greatly outperforms other contemporary models in segmentation tasks related to the aorta, left kidney, pancreas, and spleen. 
\begin{figure}[H]
    \centering
    \includegraphics[width=0.8\textwidth, height=0.3\textheight]{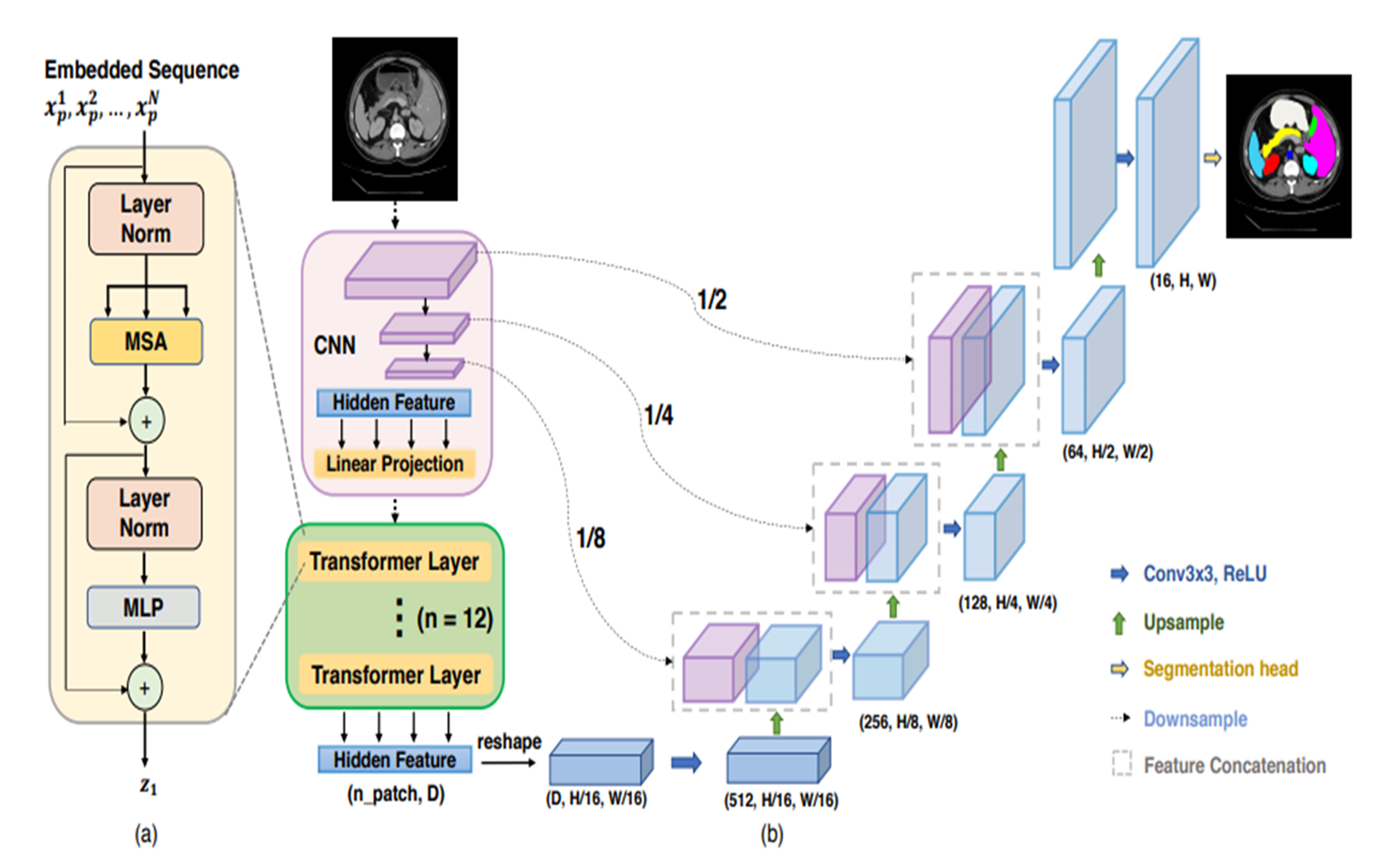}
    \caption{TransUNet Model}
\end{figure}

\hspace{0.5cm} In our experiments, we observed that TransUNet achieved a high dice score on the bone dataset, with a training and validation dice score of 0.87. 
\begin{figure}[H]
    \centering
    \includegraphics[width=1.03\textwidth, height=0.4\textheight]{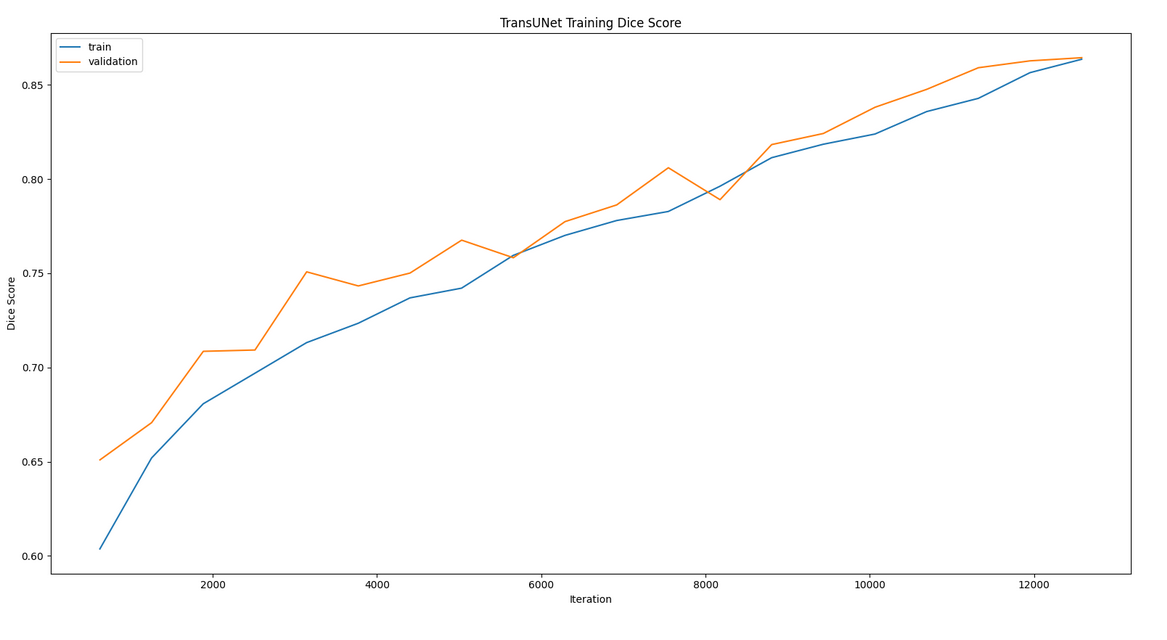}
    \caption{TransUNet Result 1}
\end{figure}

\hspace{0.5cm} However, when we expanded the dataset to also include data from the pancreas dataset, the training dice score dropped to 0.80 and the validation dice score dropped to 0.79. We speculated that this is because the TransUNet architecture is able to perform well when it is tasked with a single type of tissue and that its performance deteriorates when it has to work with a dataset that contains different types of tissues.
\begin{figure}[H]
    \centering
    \includegraphics[width=1\textwidth, height=0.4\textheight]{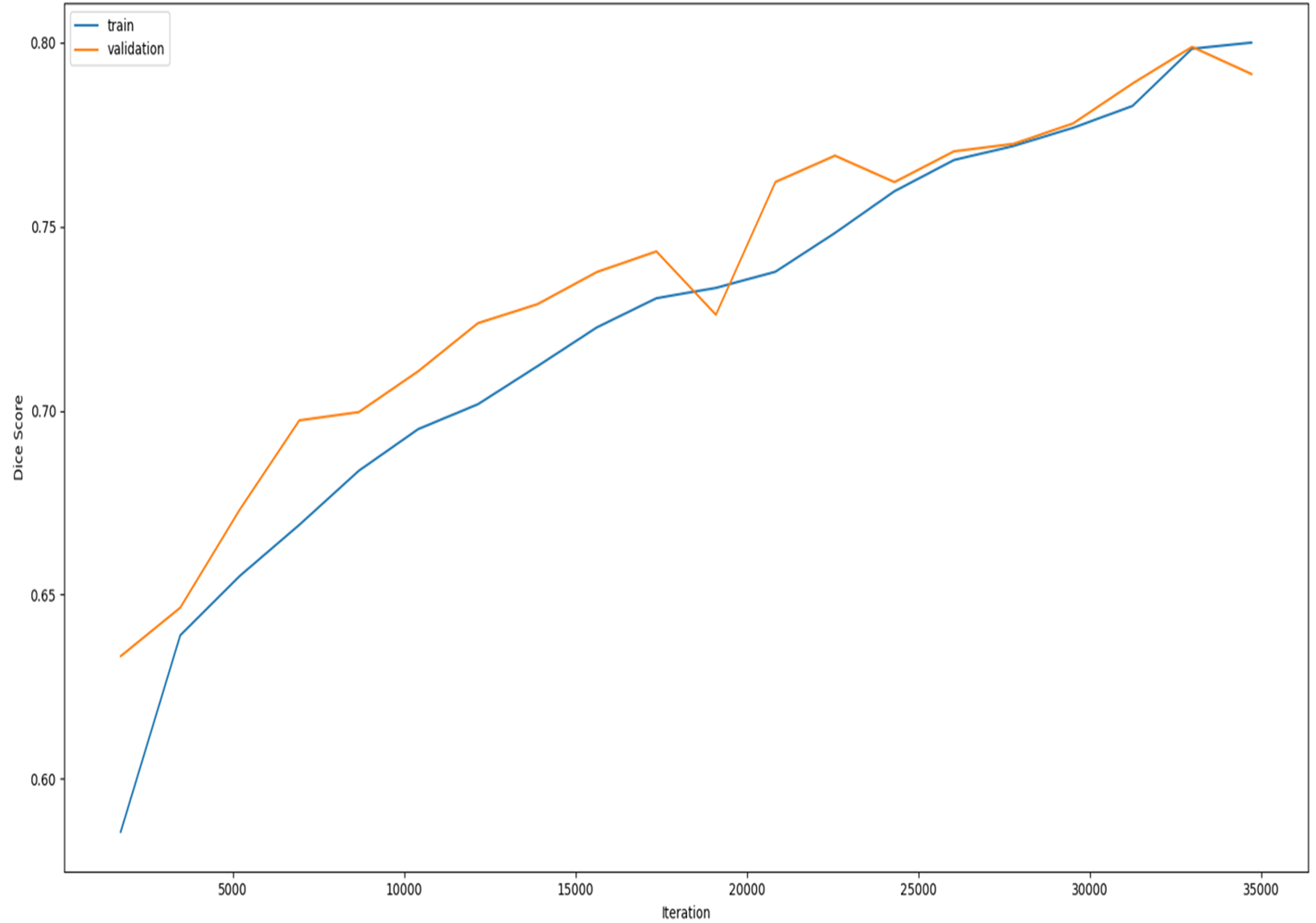}
    \caption{TransUNet Result 2}
\end{figure}

\subsection{Results}

\hspace{0.5cm} The validation results of the working models on the Bone dataset are shown below. 

\begin{figure}[H]
    \centering
    \includegraphics[width=0.7\textwidth]{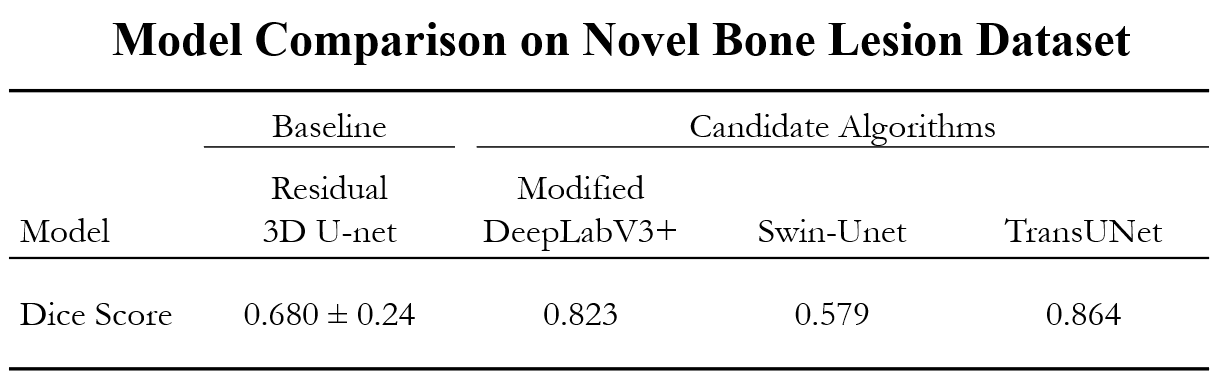}
    \caption{Model Result Comparison}
\end{figure}

\hspace{0.5cm} Based on the comparison, we decided to further train and fine-tune the TransUNet model. The last time we looked at its training curve showed its potential to generate a comparable result against the baseline. Below are some quantitative and qualitative results of TransUNet.
\begin{figure}[H]
    \centering
    \includegraphics[width=1\textwidth, height=0.4\textheight]{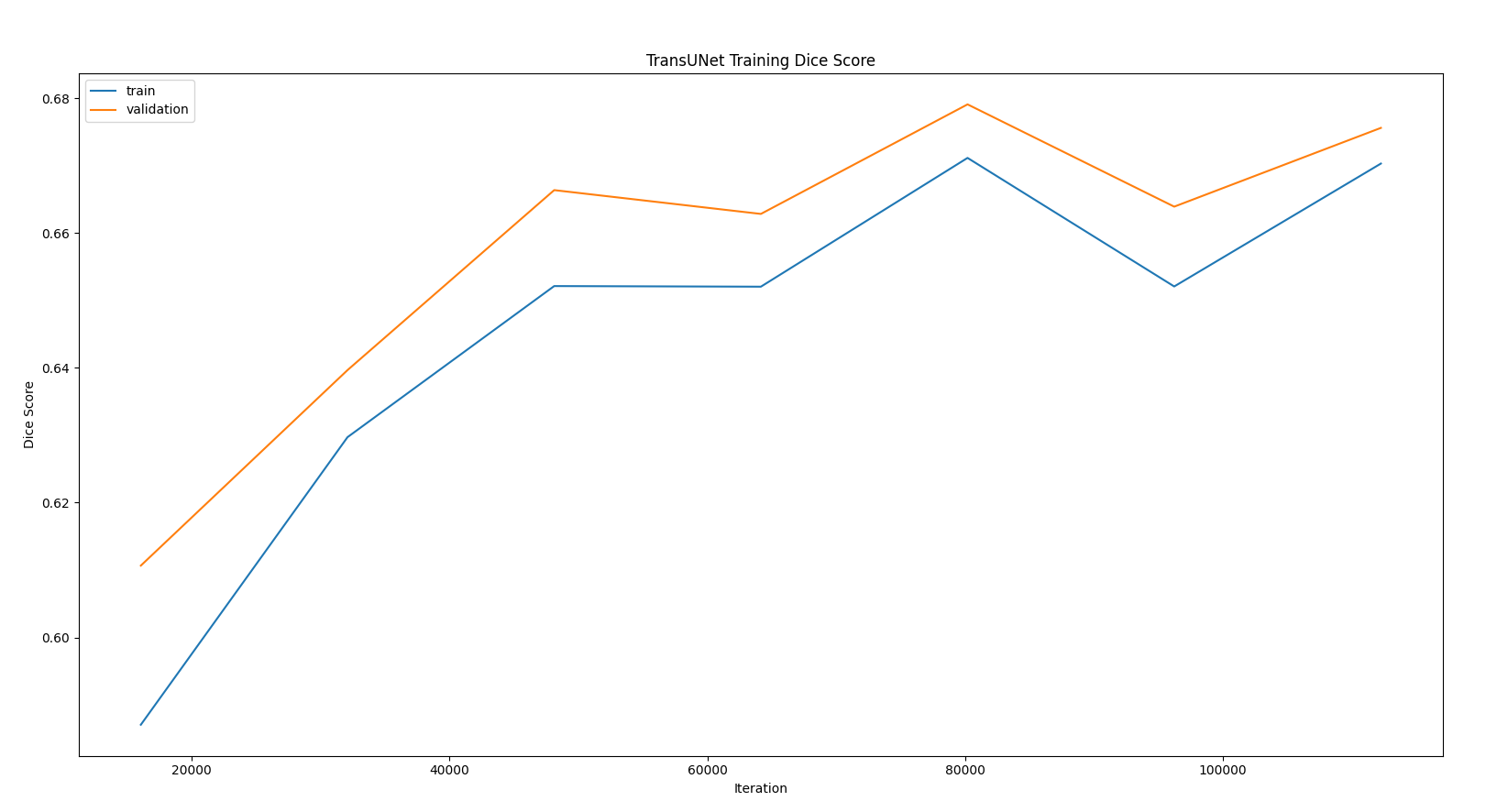}
    \caption{TransUNet Result 3}
\end{figure}

\begin{figure}[H]
    \centering
    \begin{subfigure}{0.3\textwidth}
        \includegraphics[width=\linewidth, height=0.2\textheight]{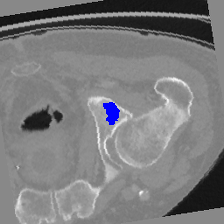}
        \caption{Ground Truth 1}
    \end{subfigure}
    \begin{subfigure}{0.3\textwidth}
        \includegraphics[width=\linewidth, height=0.2\textheight]{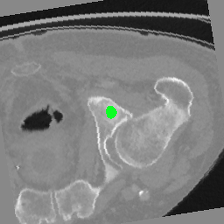}
        \caption{Inference 1}
    \end{subfigure}
    \\
    \begin{subfigure}{0.3\textwidth}
        \includegraphics[width=\linewidth, height=0.2\textheight]{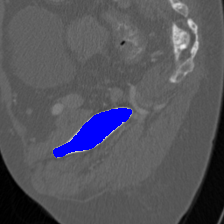}
        \caption{Ground Truth 2}
    \end{subfigure}
    \begin{subfigure}{0.3\textwidth}
        \includegraphics[width=\linewidth, height=0.2\textheight]{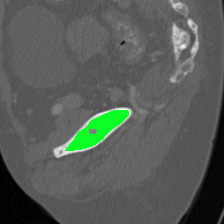}
        \caption{Inference 2}
    \end{subfigure}
    \caption{TransUNet Qualitative Results}
\end{figure}

\section{Discussion}
\hspace{0.5cm} Most of our experimented models did not work as intended and failed due to out-of-memory error. Some of them were able to run but suffered very badly when we trained our models across various parts of the dataset. In this section, we will attempt to find the reasons behind these phenomena and propose potential remedies.

\subsection{Model Size}
\hspace{0.5cm} Model sizes have been always an issue during our experimentation. Many models failed to run because the RAM requirements for their initialization were too ambitious, exceeding our current 50 GB capacity.

\hspace{0.5cm}  The reason behind this gigantic memory consumption is straightforward: the original models typically work on small data points, such as a 128x128 image; however, once we adopted it to work on the data from the ULS challenges, the immense increase in the size of the data jumped the magnitude of the parameters up, thus creating a barrier for initializing the models. 

\hspace{0.5cm} Now, there are some potential remedies. One is to combine data parallelism, pipeline partition, and model parallelism.\cite{DBLP:journals/corr/abs-1712-04432modelpara} This way the model can be split across many devices, and potentially, we can train a more complex model that can provide better results. 

\hspace{0.5cm} After getting a well-performing giant model, we can distill its knowledge to small-scale, easy-to-deploy specialist models.\cite{hinton2015distilling} After all the specialist models are trained, we can train a gated model on the giant model, distributing the data points to its corresponding specialists. This essentially composes a "mixture of experts" model\cite{MoE}, which could deviate from the original intent of the challenge host. However, it sounds to us a very effective solution to boost up the segmentation accuracy without sacrificing much of the inference speed.
\subsection{Contextual Information}
\hspace{0.5cm} Another compelling reason behind the inability of the models to perform well is, perhaps, innate to the problem at hand. One single model, with its predefined architecture, may fail to capture the variation of all the tissue types. That is to say, with the limited computation resources and strict requirements on the inference speed, it is impossible to have a universal model working well on all tissue types.

\hspace{0.5cm} To address this issue, we found Knowledge Embedding Network\cite{10.1007/978-3-031-20074-8_28KEN}, which follows dictionary learning principles to carefully select a collection of vocabularies and incorporate the context information into the inference layers using that collection. The model, trained for several epochs, was not satisfying to us concerning the training and validation loss. We encourage further investigation.

\section{Conclusion}

\hspace{0.5cm} In response to the growing demand for robust and efficient lesion segmentation models across various tissue types, a comparative research study was conducted as part of the Universal Lesion Segmentation Challenge 2023. We aimed to develop a model capable of universal lesion segmentation while maintaining fast inference times. Several state-of-the-art architectures were evaluated, including nnUNetv2, DeepLabV3+, Medical Transformer, SwinUnet, and TransUNet. Most of them did not perform well, and none of the experiments beat the baseline. Problems involved a lack of RAM in initializing models, insufficient segmentation accuracies, and etc. Based on our results, we continued training TransUNet and included a qualitative demonstration of its inferences on some slices. Finally, we discussed what could be done to potentially have a better model for the universal segmentation task.

\newpage

\bibliography{main}

\begin{thebibliography}{18}
\providecommand{\natexlab}[1]{#1}
\providecommand{\url}[1]{\texttt{#1}}
\expandafter\ifx\csname urlstyle\endcsname\relax
  \providecommand{\doi}[1]{doi: #1}\else
  \providecommand{\doi}{doi: \begingroup \urlstyle{rm}\Url}\fi

\bibitem[Cao et~al.(2021)Cao, Wang, Chen, Jiang, Zhang, Tian, and
  Wang]{cao2021swinunet}
Hu~Cao, Yueyue Wang, Joy Chen, Dongsheng Jiang, Xiaopeng Zhang, Qi~Tian, and
  Manning Wang.
\newblock Swin-unet: Unet-like pure transformer for medical image segmentation,
  2021.

\bibitem[Chen et~al.(2021)Chen, Lu, Yu, Luo, Adeli, Wang, Lu, Yuille, and
  Zhou]{DBLP:journals/corr/abs-2102-04306TransUNet}
Jieneng Chen, Yongyi Lu, Qihang Yu, Xiangde Luo, Ehsan Adeli, Yan Wang, Le~Lu,
  Alan~L. Yuille, and Yuyin Zhou.
\newblock Transunet: Transformers make strong encoders for medical image
  segmentation.
\newblock \emph{CoRR}, abs/2102.04306, 2021.
\newblock URL \url{https://arxiv.org/abs/2102.04306}.

\bibitem[Chen et~al.(2017)Chen, Papandreou, Schroff, and
  Adam]{DBLP:journals/corr/ChenPSA17DLV3}
Liang{-}Chieh Chen, George Papandreou, Florian Schroff, and Hartwig Adam.
\newblock Rethinking atrous convolution for semantic image segmentation.
\newblock \emph{CoRR}, abs/1706.05587, 2017.
\newblock URL \url{http://arxiv.org/abs/1706.05587}.

\bibitem[Chen et~al.(2018)Chen, Zhu, Papandreou, Schroff, and
  Adam]{DBLP:journals/corr/abs-1802-02611DLV3+}
Liang{-}Chieh Chen, Yukun Zhu, George Papandreou, Florian Schroff, and Hartwig
  Adam.
\newblock Encoder-decoder with atrous separable convolution for semantic image
  segmentation.
\newblock \emph{CoRR}, abs/1802.02611, 2018.
\newblock URL \url{http://arxiv.org/abs/1802.02611}.

\bibitem[{\c{C}}i{\c{c}}ek et~al.(2016){\c{C}}i{\c{c}}ek, Abdulkadir, Lienkamp,
  Brox, and Ronneberger]{DBLP:journals/corr/CicekALBR16}
{\"{O}}zg{\"{u}}n {\c{C}}i{\c{c}}ek, Ahmed Abdulkadir, Soeren~S. Lienkamp,
  Thomas Brox, and Olaf Ronneberger.
\newblock 3d u-net: Learning dense volumetric segmentation from sparse
  annotation.
\newblock \emph{CoRR}, abs/1606.06650, 2016.
\newblock URL \url{http://arxiv.org/abs/1606.06650}.

\bibitem[Dosovitskiy et~al.(2020)Dosovitskiy, Beyer, Kolesnikov, Weissenborn,
  Zhai, Unterthiner, Dehghani, Minderer, Heigold, Gelly, Uszkoreit, and
  Houlsby]{DBLP:journals/corr/abs-2010-11929Vit}
Alexey Dosovitskiy, Lucas Beyer, Alexander Kolesnikov, Dirk Weissenborn,
  Xiaohua Zhai, Thomas Unterthiner, Mostafa Dehghani, Matthias Minderer, Georg
  Heigold, Sylvain Gelly, Jakob Uszkoreit, and Neil Houlsby.
\newblock An image is worth 16x16 words: Transformers for image recognition at
  scale.
\newblock \emph{CoRR}, abs/2010.11929, 2020.
\newblock URL \url{https://arxiv.org/abs/2010.11929}.

\bibitem[Gholami et~al.(2017)Gholami, Azad, Keutzer, and
  Bulu{\c{c}}]{DBLP:journals/corr/abs-1712-04432modelpara}
Amir Gholami, Ariful Azad, Kurt Keutzer, and Aydin Bulu{\c{c}}.
\newblock Integrated model and data parallelism in training neural networks.
\newblock \emph{CoRR}, abs/1712.04432, 2017.
\newblock URL \url{http://arxiv.org/abs/1712.04432}.

\bibitem[Hinton et~al.(2015)Hinton, Vinyals, and Dean]{hinton2015distilling}
Geoffrey Hinton, Oriol Vinyals, and Jeff Dean.
\newblock Distilling the knowledge in a neural network, 2015.

\bibitem[Ho et~al.(2019)Ho, Kalchbrenner, Weissenborn, and
  Salimans]{DBLP:journals/corr/abs-1912-12180AA}
Jonathan Ho, Nal Kalchbrenner, Dirk Weissenborn, and Tim Salimans.
\newblock Axial attention in multidimensional transformers.
\newblock \emph{CoRR}, abs/1912.12180, 2019.
\newblock URL \url{http://arxiv.org/abs/1912.12180}.

\bibitem[Isensee et~al.(2020)Isensee, Jaeger, Kohl, Petersen, and
  Maier-Hein]{Isensee2020nnUNetAS}
Fabian Isensee, Paul~F. Jaeger, Simon A.~A. Kohl, Jens Petersen, and Klaus
  Maier-Hein.
\newblock nnu-net: a self-configuring method for deep learning-based biomedical
  image segmentation.
\newblock \emph{Nature Methods}, 18:\penalty0 203 -- 211, 2020.
\newblock URL \url{https://api.semanticscholar.org/CorpusID:227947847}.

\bibitem[Jacobs et~al.(1991)Jacobs, Jordan, Nowlan, and Hinton]{MoE}
R.~A. Jacobs, M.~I. Jordan, S.~J. Nowlan, and G.~E. Hinton.
\newblock Adaptive mixtures of local experts., 1991.

\bibitem[LeCun et~al.(1989)LeCun, Boser, Denker, Henderson, Howard, Hubbard,
  and Jackel]{NIPS1989_53c3bce6}
Yann LeCun, Bernhard Boser, John Denker, Donnie Henderson, R.~Howard, Wayne
  Hubbard, and Lawrence Jackel.
\newblock Handwritten digit recognition with a back-propagation network.
\newblock In D.~Touretzky, editor, \emph{Advances in Neural Information
  Processing Systems}, volume~2. Morgan-Kaufmann, 1989.
\newblock URL
  \url{https://proceedings.neurips.cc/paper_files/paper/1989/file/53c3bce66e43be4f209556518c2fcb54-Paper.pdf}.

\bibitem[Li et~al.(2018)Li, Chen, Qi, Dou, Fu, and Heng]{8379359}
Xiaomeng Li, Hao Chen, Xiaojuan Qi, Qi~Dou, Chi-Wing Fu, and Pheng-Ann Heng.
\newblock H-denseunet: Hybrid densely connected unet for liver and tumor
  segmentation from ct volumes.
\newblock \emph{IEEE Transactions on Medical Imaging}, 37\penalty0
  (12):\penalty0 2663--2674, 2018.
\newblock \doi{10.1109/TMI.2018.2845918}.

\bibitem[Milletari et~al.(2016)Milletari, Navab, and Ahmadi]{7785132}
Fausto Milletari, Nassir Navab, and Seyed-Ahmad Ahmadi.
\newblock V-net: Fully convolutional neural networks for volumetric medical
  image segmentation.
\newblock In \emph{2016 Fourth International Conference on 3D Vision (3DV)},
  pages 565--571, 2016.
\newblock \doi{10.1109/3DV.2016.79}.

\bibitem[nan Xiao et~al.(2018)nan Xiao, Lian, Luo, and Li]{Xiao2018WeightedRF}
Xiao nan Xiao, Sheng Lian, Zhiming Luo, and Shaozi Li.
\newblock Weighted res-unet for high-quality retina vessel segmentation.
\newblock \emph{2018 9th International Conference on Information Technology in
  Medicine and Education (ITME)}, pages 327--331, 2018.
\newblock URL \url{https://api.semanticscholar.org/CorpusID:57190934}.

\bibitem[Qiu and Xu(2022)]{10.1007/978-3-031-20074-8_28KEN}
Yu~Qiu and Jing Xu.
\newblock Delving into universal lesion segmentation: Method, dataset,
  and benchmark.
\newblock In Shai Avidan, Gabriel Brostow, Moustapha Ciss{\'e}, Giovanni~Maria
  Farinella, and Tal Hassner, editors, \emph{Computer Vision -- ECCV 2022},
  pages 485--503, Cham, 2022. Springer Nature Switzerland.
\newblock ISBN 978-3-031-20074-8.

\bibitem[Ronneberger et~al.(2015)Ronneberger, Fischer, and
  Brox]{RonnebergerFB15}
Olaf Ronneberger, Philipp Fischer, and Thomas Brox.
\newblock U-net: Convolutional networks for biomedical image segmentation.
\newblock \emph{CoRR}, 2015.

\bibitem[Valanarasu et~al.(2021)Valanarasu, Oza, Hacihaliloglu, and
  Patel]{DBLP:journals/corr/abs-2102-10662medtrans}
Jeya Maria~Jose Valanarasu, Poojan Oza, Ilker Hacihaliloglu, and Vishal~M.
  Patel.
\newblock Medical transformer: Gated axial-attention for medical image
  segmentation.
\newblock \emph{CoRR}, abs/2102.10662, 2021.
\newblock URL \url{https://arxiv.org/abs/2102.10662}.

\end{thebibliography}
\bibliographystyle{plainnat}

\end{document}